\documentclass[lettersize,journal]{IEEEtran}
\usepackage{amsmath,amsfonts}
\usepackage{algorithm}
\usepackage{array}
\usepackage[caption=false,font=normalsize,labelfont=sf,textfont=sf]{subfig}
\usepackage{textcomp}
\usepackage{stfloats}
\usepackage{url}
\usepackage{verbatim}
\usepackage{graphicx}
\usepackage{cite}

\usepackage{amssymb}
\usepackage{cite}
\usepackage{multirow}
\usepackage{algpseudocode}
\usepackage{booktabs}
\usepackage[table]{xcolor}
\usepackage{hyperref}

\begin{document}

\title{SIGN: Safety-Aware Image-Goal Navigation for Autonomous Drones via Reinforcement Learning}

\author{Zichen Yan, Rui Huang, Lei He, Shao Guo, Lin Zhao
\thanks{This paper is supported by the Singapore Ministry of Education Tier 2 Academic Research Funds (T2EP20123-0037,
T2EP20224-0035). (Corresponding author: Lin Zhao) The authors are with the Department of Electrical and Computer Engineering, National University of Singapore, Singapore.
        {Email: \tt \small{ zichenyan@u.nus.edu, ruihuang@u.nus.edu, lei.he@nus.edu.sg, e1374427@u.nus.edu, zhaolin@nus.edu.sg}}%
}

\thanks{Manuscript received August 18, 2025; accepted December 1, 2025.}
}

\markboth{Journal of \LaTeX\ Class Files,~Vol.~14, No.~8, August~2021}%
{Shell \MakeLowercase{\textit{et al.}}: A Sample Article Using IEEEtran.cls for IEEE Journals}


\maketitle

\begin{abstract}
Image-goal navigation (ImageNav) tasks a robot with autonomously exploring an unknown environment and reaching a location that visually matches a given target image. While prior works primarily study ImageNav for ground robots, enabling this capability for autonomous drones is substantially more challenging due to their need for high-frequency feedback control and global localization for stable flight. In this paper, we propose a novel sim-to-real framework that leverages reinforcement learning (RL) to achieve ImageNav for drones. To enhance visual representation ability, our approach trains the vision backbone with auxiliary tasks, including image perturbations and future transition prediction, which results in more effective policy training. The proposed algorithm enables end-to-end ImageNav with direct velocity control, eliminating the need for external localization. Furthermore, we integrate a depth-based safety module for real-time obstacle avoidance, allowing the drone to safely navigate in cluttered environments. Unlike most existing drone navigation methods that focus solely on reference tracking or obstacle avoidance, our framework supports comprehensive navigation behaviors, including autonomous exploration, obstacle avoidance, and image-goal seeking, without requiring explicit global mapping. Code and model checkpoints are available at \url{https://github.com/Zichen-Yan/SIGN}.
\end{abstract}


\section{Introduction}
\IEEEPARstart{L}{earning} vision-based navigation has attracted increasing attention in autonomous drone research. Existing works mostly focus on achieving autonomous obstacle avoidance and reference tracking, where the reference signal can be a target position~\cite{flying}, velocity~\cite{hu2025seeing, bhattacharya2025vision, davide1} or direction~\cite{Navrl, davide3}.
ImageNav poses a further challenge: enabling autonomous drones to search an unknown environment for a given goal image while avoiding collisions. This problem remains largely unexplored but has significant value in practical applications such as disaster rescue and industrial inspection, especially when pre-built maps and external localization are unavailable. Purely vision-based navigation reduces the dependence on additional sensors to enable a low-cost and lightweight drone design. In this paper, we investigate ImageNav for autonomous drones using only visual and IMU measurements.

The body of ImageNav research has primarily focused on ground robots~\cite{zer, fgp, ovrlv2, zson}. Most existing studies are conducted exclusively in simulators such as Habitat~\cite{habitat2} and neglect to penalize collisions with obstacles, which can result in unsafe behaviors when deployed in the real world. Meanwhile, discrete action spaces such as \{\textit{move\_forward}, \textit{turn\_left}, \textit{turn\_right}, \textit{stop}\} are commonly adopted to simplify learning by reducing the search space. This is generally only acceptable for ground robots, which are inherently stable and can correct their position through vision by the discrete low-frequency actions.

In contrast, achieving safe and efficient ImageNav for autonomous drones is considerably more challenging. Drones are inherently unstable platforms with fast dynamics, necessitating high-frequency feedback control and accurate global localization to maintain stable flight. When relying on visual-inertial localization, they are particularly susceptible to pose-estimation drift, especially in indoor environments. Furthermore, discrete-action policies often lack the precision and responsiveness needed for rapid pose correction. 
For flight control systems, frequent switching between discrete actions can severely degrade real-time tracking performance due to the abrupt change of references. In this case, a continuous policy that directly outputs velocity commands can address this issue by enabling smoother and more timely control. Continuous actions also support seamless integration with flight controllers without additional planning modules. However, it also expands the exploration space, substantially increasing training complexity. In addition, naively embedding safety constraints into policy learning can hinder effective exploration and degrade navigation performance, particularly in cluttered or narrow environments such as Gibson~\cite{gibson}, MP3D~\cite{mp3d}, and HM3D~\cite{hm3d}. Therefore, balancing efficient exploration with robust real-world safety remains a central challenge in autonomous drone navigation.

\begin{figure*}[!t]
    \centering
    \includegraphics[width=0.9\linewidth]{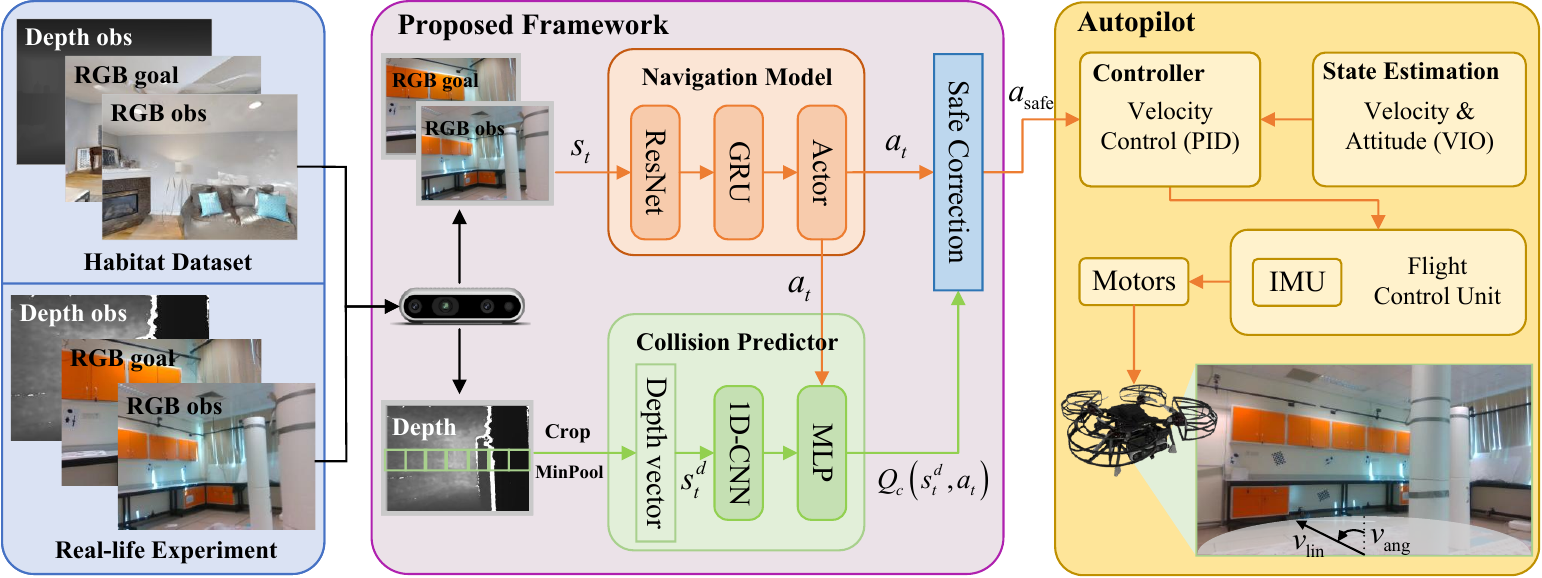}
    \caption{SIGN enables end-to-end ImageNav for autonomous drones via visual reinforcement learning with effective sim-to-real transfer. Guided by the collision predictor, an action correction mechanism is incorporated to ensure obstacle-avoiding navigation in indoor environments.}
    \label{fig:title}
\end{figure*}

In this paper, we introduce \textbf{SIGN} (\textbf{S}afety-aware \textbf{I}mage-\textbf{G}oal \textbf{N}avigation), an end-to-end RL-based ImageNav framework for autonomous drones that employs continuous velocity control and incorporates a safety module for collision avoidance. Our main contributions are as follows:
\begin{itemize}
    \item SIGN performs end-to-end ImageNav on real drones using continuous velocity control, with only visual and IMU inputs and does not rely on external localization. It supports seamless sim-to-real transfer without any finetuning and is validated on both simulation benchmarks and real-world experiments.

    \item We propose self-supervised auxiliary tasks that significantly improve training efficiency and achieve state-of-the-art performance on the Gibson benchmark under continuous control. These tasks can be seamlessly integrated into on-policy RL training. More broadly, SIGN offers a general framework for extending existing discrete-action ImageNav algorithms to continuous control settings.

    \item Moreover, while existing end-to-end ImageNav algorithms often fail to account for safety, we introduce a safety module that predicts collision probability from depth inputs and corrects potentially unsafe actions, thereby ensuring reliable obstacle avoidance.
\end{itemize}

\section{Related Works}
\subsection{Visual Navigation on Drones}

We focus on end-to-end, map-less vision-based navigation without relying on global localization. Prior research in this direction has mainly addressed reference-tracking control with obstacle avoidance~\cite{flying, hu2025seeing, davide1, bhattacharya2025vision, Navrl} or drone racing on a fixed course~\cite{davide2}, where an RL policy maps sensor inputs directly to control commands such as velocity or acceleration. 
The key distinction from these tasks is that ImageNav specifies the reference signal through an image, which serves as a visual goal for navigation. Similarly, Tao~\cite{tao2024learning} explored RL-based indoor search using drones, but their approach depends on global mapping. In contrast, our task is more challenging as it requires not only map-less exploration and collision avoidance, but also navigation toward a specified goal image.

\subsection{End-to-end RL-based Visual Navigation}
For visual navigation tasks, end-to-end RL has demonstrated higher efficiency in real-time inference and module tuning by avoiding intermediate stages such as mapping and subgoal selection~\cite{neuralslam}. Early work by Wijmans~\cite{ddppo} addressed point-goal visual navigation for ground robots. However, it relied on GPS and compass data, which is impractical for indoor settings. Al-Halah~\cite{zer} removed the dependency on localization and proposed a map-less visual navigation framework, enabling zero-shot generalization to new tasks and multi-modal goals. Additionally, previous works have employed memory mechanisms~\cite{memaug} and topological maps~\cite{vgm, tsgm, DBLP:conf/nips/HahnCTMRG21} to enhance exploration efficiency by storing keyframes and leveraging graph neural networks to extract visual correlations for guidance. For now, most existing end-to-end RL methods are constrained to discrete actions and suffer from training inefficiency due to environmental diversity. To better facilitate drone deployment and simplify the sim-to-real transfer, we investigate the extension from discrete to continuous settings.

\subsection{Image-conditioned Waypoint Planner}
Many works predict waypoints to reach a target image and use low-level position controllers for trajectory tracking~\cite{nomad, vint, gnm}, even assisted by a world model~\cite{ bar2025navigation}. A common weakness of these methods lies in the out-of-distribution (OOD) issues inherent to the supervised learning paradigm, especially when the current and target images have no overlap. In many image-conditioned navigation methods, continuous policies have already been explored through imitation learning. However, ImageNav involves numerous OOD scenarios that make generalization challenging without global localization to guide exploration.
To address this issue, an RL-based continuous policy presents a promising solution.
Additionally, an emerging trend is to combine 3D reconstruction techniques with ImageNav for drones, such as NeRF~\cite{adamkiewicz2022vision} and Gaussian Splatting~\cite{SINGER}. They typically require offline reconstruction and substantial computational resources. Due to the limited onboard capacity of drones, real-time processing of such methods remains impractical.
  
\begin{figure}[!b]
\vspace{-1em}
    \centering
    \includegraphics[width=0.75\linewidth]{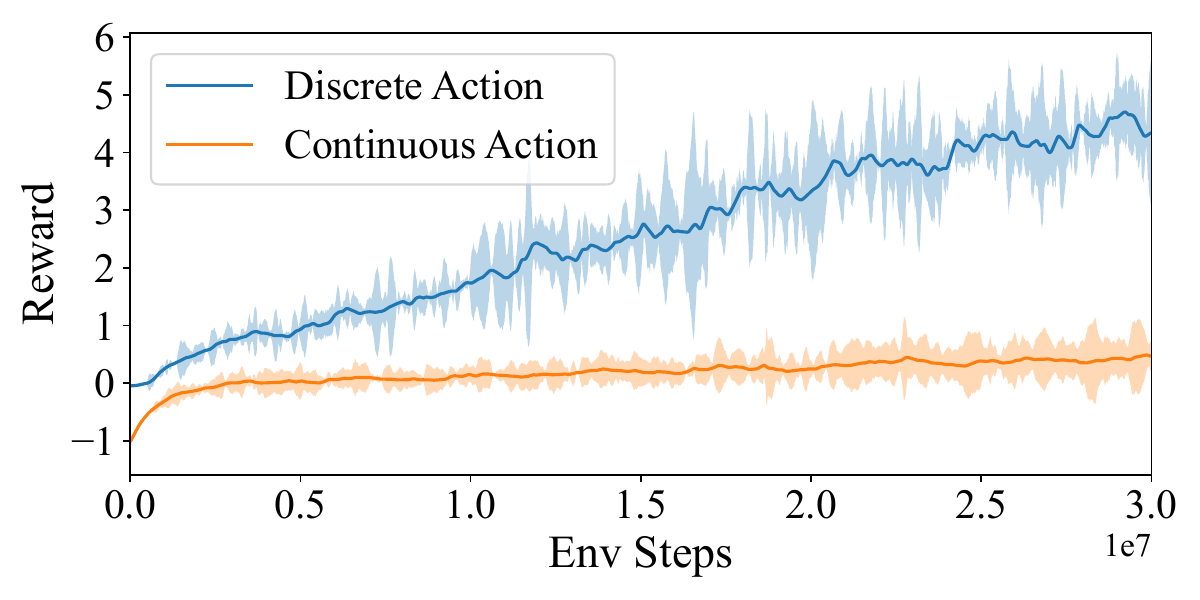}  
    \caption{Performance comparison between discrete and continuous action using the SoTA baseline FGPrompt \cite{fgp}.}
    \label{fig:comparison}
\end{figure} 

\section{Visual Reinforcement Learning for ImageNav}
As depicted in Fig.~\ref{fig:title}, SIGN is an end-to-end framework for drone ImageNav that leverages a continuous policy to enable map-less navigation with enhanced safety. But training continuous policies for ImageNav presents a significant challenge due to the expanded action space and increased training complexity. In Fig.~\ref{fig:comparison}, we compare a discrete State-of-The-Art (SoTA) baseline with its continuous counterpart, where the discrete action space $\mathcal{A}_d=\{\textit{move\_forward}, \textit{turn\_left}, \textit{turn\_right}, \textit{stop}\}$ and the continuous one consists of forward linear and angular velocities, $\mathcal{A}_c=\{v_{lin},v_{ang}\}$. It can be observed that the discrete-action policy achieves steadily increasing rewards and much higher training efficiency, while the continuous-action policy suffers from very slow reward growth. This highlights the difficulty of training continuous policies in ImageNav tasks and motivates the design of our SIGN framework to address these challenges.

\subsection{Task Definition} 
Given a goal RGB image in an unknown environment, the agent is required to search the area for the target based on first-person observations from a single camera.
At each step $t$, the agent samples an action $a_t$ conditioned on visual observations $s_t$, and gets a reward $r_t$. The task is considered successful if the distance to the goal is less than $1~m$ and the angle between the target image and current image is less than $25^\circ$.

\subsection{Navigation Model}
\begin{figure}[!t]
\vspace{+0.5em}
    \centering
    \includegraphics[width=\linewidth]{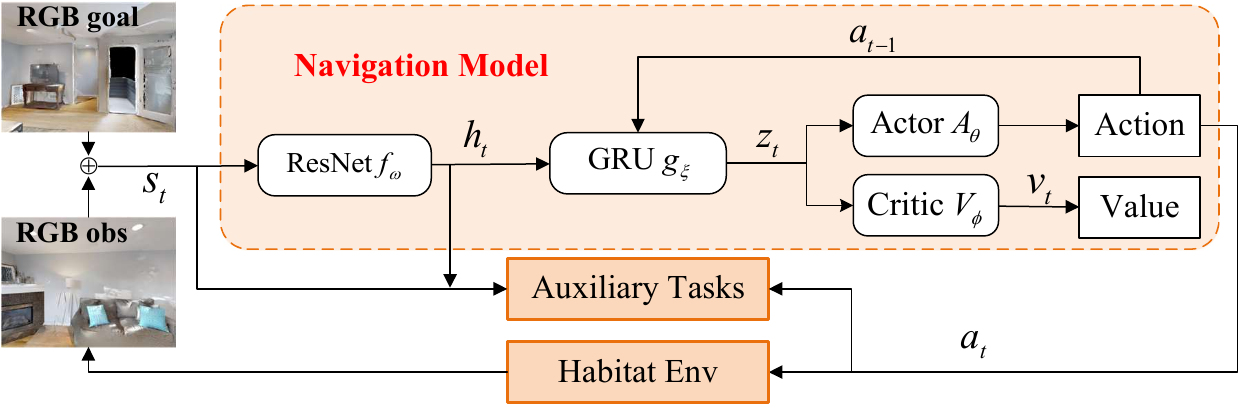}
    \caption{Overview of the navigation model.}
    \label{fig:overview}   
    \vspace{-1em}
\end{figure}

The navigation model consists of a ResNet encoder $f_{\omega}$, a GRU model $g_{\xi}$, and an actor-critic policy trained by Proximal Policy Optimization (PPO) \cite{ppo}. See Fig.~\ref{fig:overview} for the diagram. The goal image $O_g$ and current image $O_c$ are stacked as the input $s_t$. The inference procedure can be summarized as 
\begin{equation}
\begin{aligned}
    s_t & =O_c \oplus O_g, \\
    h_t &= f_{\omega}(s_t), z_t = g_{\xi}(h_t, a_{t-1}),\\
    a_t &\sim A_{\theta}(\cdot | z_t), v_t = V_{\phi}(z_t).
\end{aligned}
\end{equation}

\subsection{Augmented Continuous Policy Training}

A possible reason of the inefficient continuous policy training in Fig.~\ref{fig:comparison} is that, in the early stage, training an effective vision encoder from scratch is time-consuming. RL policy struggles to improve until the vision encoder learns to capture key information from pixels. To address this problem, we propose two useful Self-Supervised Learning (SSL) techniques as auxiliary tasks to enhance the policy learning. 

Recall the standard optimization form in vanilla PPO \cite{ppo}
\begin{equation}
\begin{aligned}
    &\max_{\theta} L_a(\theta) = \mathbb{E}_t \left[ \min (r_t(\theta)\hat{A}_t,\text{clip}(r_t(\theta), 1-\epsilon,1+\epsilon)\hat{A}_t)\right]\\
    &\min_{\phi} L_v(\phi) = \mathbb{E}_t \left[\frac{1}{2} (V_{gt}- V_{\phi}(s_t))^2\right]
\end{aligned}
\end{equation}
where $\hat{A}_t$ adopts generalized advantage estimation and $r_t(\theta) = {\pi_{\theta}(a_t|s_t)}/{\pi_{\theta_{old}}(a_t|s_t)}$. Auxiliary tasks are built on the PPO structure and aim to robustify policy and enhance representation ability by learning to predict future states.

\begin{figure}[!b]
\vspace{-1em}
    \centering
    \includegraphics[width=\linewidth]{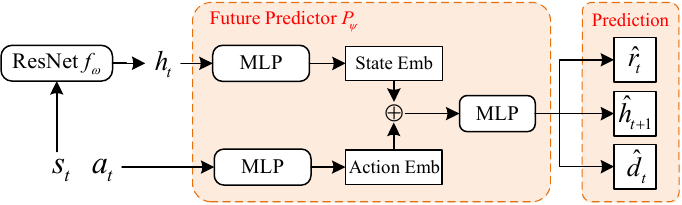}
    \caption{Auxiliary task to enhance the vision encoder $f_{\omega}$ by predicting the reward, future latent state, and done flag.}
    \label{fig:mrq}
\end{figure}

\subsubsection{Future Prediction} The prediction ability receives widespread attention due to its success in the model-based visual RL research~\cite{dreamerv3, bar2025navigation}, where a foundation world model acts as an internal simulator to imagine future observations and guide policy learning. Focusing on a portable model for drone applications, here we train a single-step prediction to enhance the visual backbone training. To this end, future prediction task is introduced, as illustrated in Fig.~\ref{fig:mrq}. We deploy an MLP head $P_{\psi}$ after the visual encoder $f_{\omega}$ to predict the reward $r_t$, the next-step latent feature $h_{t+1}=f_{\omega}(s_{t+1})$, and the done flag $d_t$.
The auxiliary training loss is defined as 
\begin{equation}
\begin{aligned}
    L_{fp}(\omega, \psi) &= \lambda_r \underbrace{\text{CrossEntropy}(\hat{r}_t, \text{Two-hot}(r_t))}_{\text {reward loss}} \\ &+ \lambda_d \underbrace{\text{MSE}(\hat{h}_{t+1}, h_{t+1})}_{\text {dynamic loss}}  + \lambda_T \underbrace{\text{MSE}(\hat{d}_t, d_t)}_{\text {termination loss}}.
\end{aligned}
\label{eq:fp} 
\end{equation}

The forward prediction of dynamics $\hat{h}_{t+1} = P_{\psi}(a_t,h_t)$ is implemented in the latent space to avoid deconvolution operations and save computation resources. The reward loss function uses cross-entropy between the predicted reward $\hat{r}_t$ and a two-hot encoding of reward $r_t$, which proves more effective in predicting sparse rewards and robust to reward magnitude \cite{mrq}.

\subsubsection{RandomShift}
The RandomShift \cite{DBLP:conf/iclr/YaratsFLP22} technique performs data augmentation on the current image $O_c$ in each sample batch. Denote the perturbated data as $aug(s_t)= randomshift(O_c) \oplus O_g$, then auxiliary loss is defined as 
\begin{equation} 
\begin{aligned}
        L_{rs}(\theta,\phi) &= \text{KL}(\underbrace{\pi_{\theta}(\cdot|s_t)}_{\text{detach}}, \pi_{\theta}(\cdot|aug(s_t))) \\
        &+\text{MSE} (\underbrace{V_{\phi}(s_t)}_{\text{detach}}, V_{\phi}(aug(s_t))) ,
\end{aligned}
\label{eq:randomshift}
\end{equation}
where \text{`detach'} operation is applied to block gradients.
The purpose of RandomShift is to maintain the action distribution and value prediction when dealing with slightly shifted images. It helps improve the model's robustness in the presence of jitter in real-life images.
The goal image $O_g$ remains unchanged in auxiliary training to align with the reward distribution. Finally, the actor-critic network can be updated with RandomShift. By adding Eq.~\eqref{eq:randomshift} into the PPO update rule, the augmented optimization becomes 
\begin{equation} \label{eq:all}
    \min_{\theta, \phi} -L_a+L_v+\lambda_{rs}L_{rs}
\end{equation}
The whole training process is shown in Algorithm \ref{algo:1}. 

\begin{algorithm}[!t]
\caption{Augmented Continuous Policy Training\label{algo:1}}
\begin{algorithmic}[1]
\State \textbf{Input:} number of env $B$, trajectory length $L$, fixed weights $\lambda_r,\lambda_d,\lambda_T, \lambda_{rs}$
\State Initialize actor-critic models $A_{\theta}$ and $V_{\phi}$ including the backbone $f_{\omega}$ and GRU $g_{\xi}$
\State Initialize predictor model $P_{\psi}$ and collision predictor $Q_c$
\For{each epoch}
    \State Collect one trajectory $\{s_t,a_t,s_{t+1},r_t, c_t,d_t\}_{1:L}$ from each env, a total of $B \times L$ samples
    
    \# \textit{Future Prediction}
    \For{each mini-batch}
        \State Update $P_{\psi}$ and $f_{\omega}$ with data $\{s_t,a_t,s_{t+1},r_t,d_t\}$ via minimizing Eq.~\eqref{eq:fp}
    \EndFor
    
    \# \textit{PPO training with RandomShift}
    \For{each mini-batch}
        \State Update policy parameters $\omega, \xi, \theta, \phi$ via Eq.~\eqref{eq:all}
        \State Train collision predictor $Q_c$ via minimizing Eq.~\eqref{eq:cost}
    \EndFor
\EndFor
\end{algorithmic}
\end{algorithm}

\section{Safety Module for Real-world Experiments} \label{chapter:4}

Existing ImageNav algorithms primarily focus on navigation performance while neglecting safety. Navigation policies trained in Habitat in those works are allowed to collide with obstacles. In practice, enforcing safety during policy training can lead to conservative behavior, which significantly hinders exploration in narrow indoor environments. Inspired by the safety filter designs in the fields of safe RL and control~\cite{usl, wabersich2021predictive, brunke2022safe}, we propose a practical action correction mechanism that is activated only when the agent is in danger. This decoupled design is shown to be effective in balancing exploration and safety. Meanwhile, it also simplifies training.

\subsection{Model Training for Collision Probability Prediction}
The depth image is first preprocessed by cropping and min-pooling operations to obtain a depth vector $s^d_t \in \mathbb{R}^n$, indicating the closest distance in different directions.
$Q_c$ is used to predict the collision probability based on the current depth image and navigation action $a_t$, defined as
\begin{equation}
        Q_c(s^d_t,a_t)\!=\mathbb{E}_t [c_{t}|s^d_t,a_t], \quad
c_t = \begin{cases}
    1, & \text{if collision,} \\
    0, & \text{else}.
\end{cases}
\end{equation}
where $c_t$ is a collision feedback from the Habitat simulator. 

However, our experimental results show that using only hard labels $c_t$ leads to poor prediction, as the predicted probability does not vary smoothly with distance and lacks a gradual transition from safe to dangerous states. To obtain more reliable probability estimates, we introduce soft labels $c_t^{\text{soft}}$ that provide a smoother and more accurate indication of collision likelihood:
\begin{equation}\label{eq:soft}
c_t^{\text{soft}} = \text{Sigmoid}\left(10 \cdot (\beta - \min_i s^d_t(i))\right),
\end{equation}
where $\min(\cdot)$ denotes the minimum element of vector $s_t^d$ and $\beta$ is the threshold for a safe distance. When the minimum perceived distance falls below $\beta$, the soft label value increases rapidly toward 1, and it decreases to 0 when the distance exceeds $\beta$. Thus, collision detection is defined with respect to a safety margin distance of $\beta$.

The collision predictor $Q_c$ is trained using a 1:1 mixture of hard and soft labels. The final loss function $L_c$ becomes 
\begin{align}
    \text{Label}_{\text{mix}} &= \text{Mix}(c_t, c_t^{soft}),\\
    L_c &= \text{BCELoss}(\hat{Q}_c(s^d_t,a_t), \text{Label}_{\text{mix}}). \label{eq:cost}
\end{align}
The benefits of mixed labels are two-fold: 1) Soft labels $c_t^{soft}$ enable smooth predictions since they contain continuous depth information; 2) Hard labels help to learn the correlation between action $a_t$ and the collision probability $\hat{Q}_c(s^d_t,a_t)$, facilitating the following action correction mechanism. As is shown in Algorithm \ref{algo:1}, $Q_c$ can be trained jointly with the RL policy in the Habitat simulator.

\subsection{Action Correction for Obstacle Avoidance}
\begin{figure}[!t]
    \centering
    \includegraphics[width=0.95\linewidth]{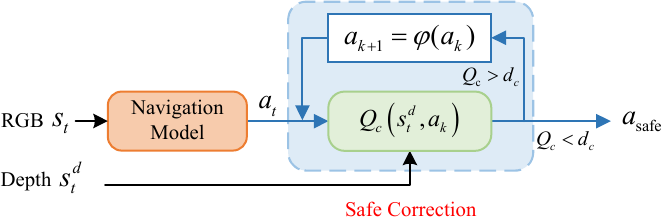}
    \caption{A correction mechanism for navigation actions to enhance safety.}
    \label{fig:correct}
    \vspace{-1em}
\end{figure}

The predicted collision probability is conditioned on $s_t^d$ and $a_t$, which means we can modify the action to reduce the collision probability. 
The correction mechanism is illustrated in Fig.~\ref{fig:correct}.
When the value of $Q_c$ exceeds a pre-defined threshold $d_c$, it indicates a dangerous situation, then actions from the navigation policy will be corrected iteratively until $Q_c<d_c$. The correction function $\varphi(\cdot)$ will project the raw action $a_0 = a_t$ to a safer one based on gradients \cite{usl}
\begin{equation}\label{USL}
    a_{k+1}=\varphi(a_{k}) = a_k- \frac{\partial}{\partial a_k} [Q_c(s_t^d, a_k)-d_c]^+,
\end{equation}
where $k=0,1,\ldots,n$ is the iteration number. The adjustment of the probability threshold $d_c$ can be viewed as a trade-off between safety and navigation performance. In general cases, $d_c$ is set to 0.5 because when $Q_c>0.5$, the minimum distance is less than the safety distance $\beta$ according to Eq.~\eqref{eq:soft}.  

\section{Experiments}
This section aims to answer the following key questions:
\begin{itemize}
    \item Can SIGN enable efficient training of continuous ImageNav policies? Do the proposed auxiliary tasks contribute to increased sample efficiency?
    \item How to achieve sim-to-real transfer from the Habitat simulator to real drones?
    \item Is the safety module able to ensure obstacle avoidance during visual navigation?
\end{itemize}
\subsection{Simulation Settings}
\subsubsection{Datasets} 
Our navigation model is trained on Gibson ImageNav dataset \cite{gibson}, including 72 training scenes and 14 testing scenes. To further validate the robustness of our algorithm, the trained model is also evaluated on cross-domain datasets MP3D \cite{mp3d} and HM3D \cite{hm3d}. Scenes from different datasets have varying backgrounds and layouts but are all simulated in the Habitat. 

\subsubsection{Agent Configurations}
To be consistent with SoTA works, we employ the Habitat camera that has a $90^\circ$ Field of View (FoV) and $128 \times 128$ resolution. The continuous action space $\mathcal{A}_c$ consists of forward linear velocity $v_{lin} \in [0,0.25]~m/s$ and angular velocity $v_{ang} \in [-15, 15]~^\circ/s$. To implement the \textit{`Stop'} action in a continuous setting, the episode will be terminated early when $v_{lin}<0.025~m/s \ \& \ |v_{ang}|<1.5~^\circ/s$.

\begin{table}[!b]
\vspace{-1em}
    \caption{Comparison with SoTA methods on Gibson.\label{table:gibson}}
    \centering
    \setlength{\tabcolsep}{3pt}
    \begin{tabular}{ccccccc}
    \hline Method & Backbone & Sensor(s) & Action Space & SPL($\uparrow$) & SR($\uparrow$) \\
    \hline 
    Mem-Aug \cite{memaug} & ResNet18  & 4 RGB & Discrete & 56.0\% & 69.0\% \\
    VGM \cite{vgm} & ResNet18  & 4 RGB & Discrete & 64.0\% & 76.0\% \\
    TSGM \cite{tsgm} & ResNet18  & 4 RGB & Discrete & 67.2\% & 81.1\% \\
    \hline
    ZER \cite{zer} & ResNet9  & RGB & Discrete & 21.6\% & 29.2\% \\
    ZSON \cite{zson} & ResNet50  & RGB & Discrete & 28.0\% & 36.9\% \\
    OVRL-V2 \cite{ovrlv2} & ViT-Base  & RGB & Discrete &58.7\% & 82.0\% \\
    FGPrompt \cite{fgp} & ResNet9  & RGB & Discrete & 66.5\% & 90.4\% \\
    \hline
    FGPrompt \cite{fgp} & ResNet9  & RGB & Continuous & 35.0\% & 79.9\% \\
    \rowcolor{gray!20}
    SIGN (Ours) & ResNet9 & RGB & Continuous &  53.3\% & 86.3\% \\
    \hline
    \end{tabular}
\end{table}

\begin{table}[!b]
\vspace{-1em}
    \caption{Cross-domain evaluation on MP3D and HM3D.\label{table:cross}}
    \centering
    \setlength{\tabcolsep}{3pt}
    \begin{tabular}{lccccc}
    \hline \multirow{2}{*}{ Methods } & \multirow{2}{*}{ Action Space } & \multicolumn{2}{c}{ MP3D } & \multicolumn{2}{c}{ HM3D } \\
    \cline { 3 - 6 } & & SPL($\uparrow$) & SR($\uparrow$) & SPL($\uparrow$) & SR($\uparrow$) \\
    \hline Mem-Aug \cite{memaug} & Discrete& $3.9\% $ & $6.9\% $ & $3.5\% $ & $1.9\% $ \\
    NRNS \cite{DBLP:conf/nips/HahnCTMRG21}& Discrete & $5.2\% $ & $9.3\% $ & $4.3\% $ & $6.6\% $ \\
    ZER \cite{zer} & Discrete& $10.8\% $ & $14.6\% $ & $6.3\% $ & $9.6\% $ \\
    FGPrompt \cite{fgp} & Discrete& $5 0 . 4\% $ & $7 7 . 6\% $ & $4 9 . 6\% $ & $7 6 . 1\% $ \\
    \hline
    FGPrompt \cite{fgp} & Continuous& $22.9\% $ & $59.4\% $ & $20.1\% $ & $56.5\% $ \\
    \rowcolor{gray!20}
    SIGN (Ours) & Continuous& $35.0\%$ & $65.4\% $ & $30.9\% $ & $64.5\%$ \\
    \hline
    \end{tabular}
\end{table}

\subsubsection{Reward and Performance Metrics}
The design of the reward function $r_t$ follows ZER \cite{zer}, which encourages the policy to approach the target and reduce the angle between the current view and the view of the goal. 
\begin{equation}
    r_t=r_{dis}(d_t,d_{t-1})+[d_t \leq d_s][\alpha_t \leq \alpha_s]r_{suc}+\gamma,
\end{equation}
where $d_t$ denotes the distance to the target position at time $t$ and $r_{dis}(d_t,d_{t-1})=d_{t-1}-d_t$ is the approaching reward. $\alpha_t$ indicates the angle between the current view and the target view. $[\cdot]$ is defined as an indicator function 
\begin{equation}
    [x] = 
    \begin{cases}
        1, & \text{if} \ x \ \text{is True}, \\
        0,  & \text{otherwise}.
    \end{cases}
\end{equation}

In our experiments, $d_s=1~m$ and $\alpha_s=25^\circ$ denote the success thresholds for distance and angular deviation, respectively. $r_{suc}=2.5$ is the success reward and $\gamma=-0.01$ is the slack penalty. On benchmarks, Success Rate (SR) and Success Weighted by Path Length (SPL) are two main metrics to quantify the navigation performance. SPL is calculated as $\frac{1}{N}\Sigma^{N}_{i=1}S_i\frac{L_i}{max(L_i,p_i)}$, where $N$ is the total number of episodes, $S_i\in \{0, 1\}$ is a success indicator, $p_i$ is the actual path length in the $i$-th episode, and $L_i$ is the optimal path length from the start to the goal point.

\subsubsection{Training Details}
To enable fast inference and hardware-friendly deployment, SIGN uses a lightweight backbone ResNet9. The whole model contains 1.6M parameters and is trained from scratch for 300M steps using an RTX 4090 GPU. PPO adopts $B=8$ parallel environments for distributed training on the Gibson benchmark. For the rest of the hyperparameters in Algorithm \ref{algo:1}, we choose $\lambda_d=1, \lambda_r=0.1, \lambda_T=0.1, \lambda_{rs}=0.5$ (loss weights, see Eq.\eqref{eq:fp} and~\eqref{eq:all}), and $L=256$.

\begin{figure}[!t]
\vspace{+0.5em}
    \centering
    \includegraphics[width=0.78\linewidth]{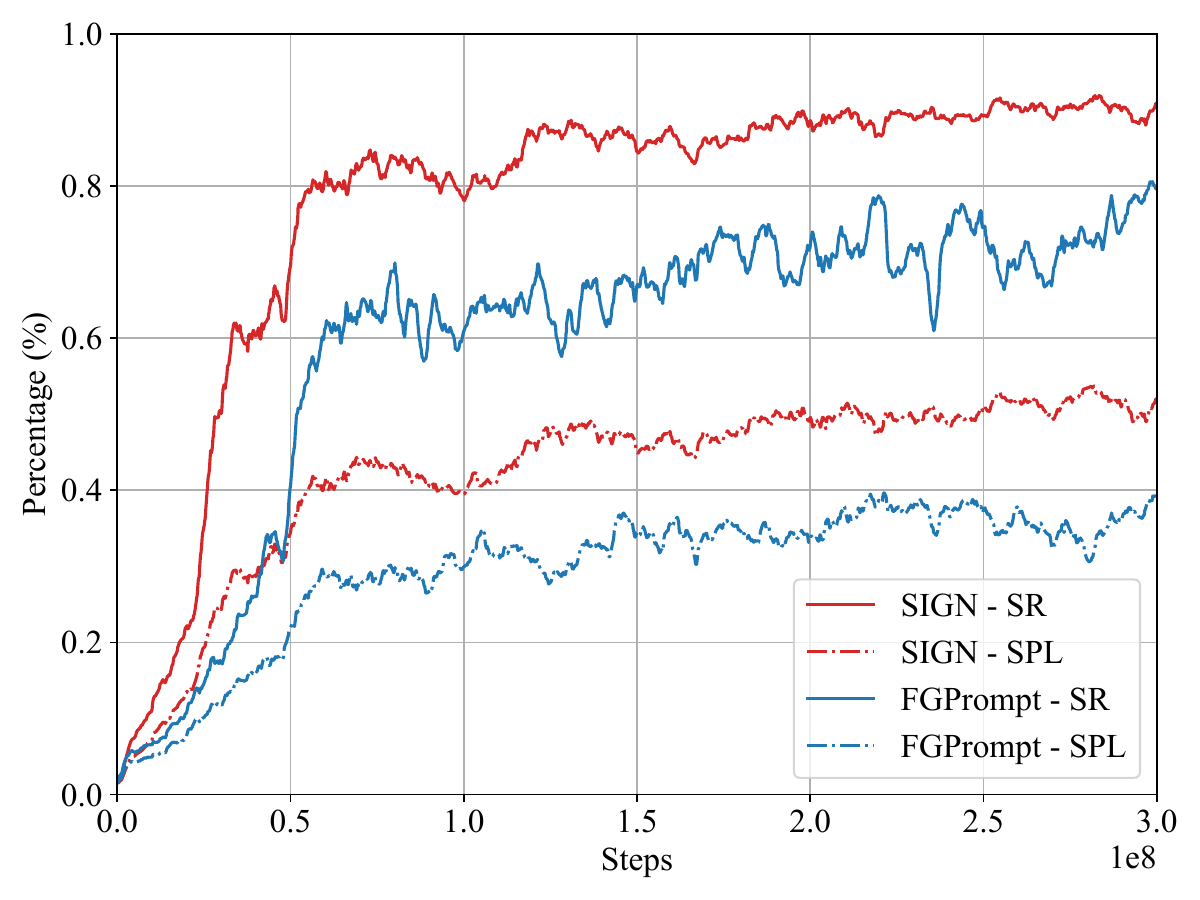}
    \caption{Comparison of training curves between SIGN and the continuous baseline.}
    \label{fig:cts_compare}
    \vspace{-1em}
\end{figure}

\subsection{Performance Analysis on Benchmarks}
\subsubsection{Training Results}
The comparison between SIGN and other SoTA algorithms is presented in Table~\ref{table:gibson}. Due to the lack of existing continuous baselines, we extend FGPrompt (a discrete SoTA method) to a continuous policy by modifying its action layer. To ensure a fair comparison, both continuous FGPrompt and SIGN are trained for the same number of steps, as illustrated in Fig.~\ref{fig:cts_compare}. On the Gibson benchmark, SIGN outperforms the continuous baseline by \textbf{+6.4\%} in SR and \textbf{+18.3\%} in SPL.
With the augmentation of the proposed auxiliary tasks, SIGN demonstrates better training efficiency than the baseline. 

The performance gap between discrete and continuous policies arises from the increased task complexity. Specifically, discrete policies simplify system dynamics by updating the agent’s pose through teleportation. In contrast, continuous velocity control follows first-order integral dynamics $(\dot{p}=v)$, requiring the agent to reason over continuous transitions and jointly optimize linear and angular velocities. As a result, training continuous-control policies is more challenging for RL, especially for early exploration and policy convergence. The significant performance degradation of FGPrompt when transferred from discrete to continuous actions further highlights the challenges of continuous control in ImageNav.

To further evaluate the generalization ability, algorithms trained on Gibson are directly transferred to unseen scenarios on MP3D and HM3D datasets. The results are averaged across easy, medium, and hard modes. Each mode contains 1000 episodes. 
The evaluation results on the cross-domain datasets are shown in Table~\ref{table:cross}. SIGN outperforms continuous FGPrompt by \textbf{+6\%} SR and \textbf{+12.1\%} SPL on MP3D, and \textbf{+8\%} SR and \textbf{+10.8\%} SPL on HM3D. Overall, SIGN is thoroughly validated across diverse ImageNav benchmarks and achieves SoTA in the continuous setting.

\begin{figure}[!t]
\vspace{+0.5em}
    \centering
    \includegraphics[width=0.75\linewidth]{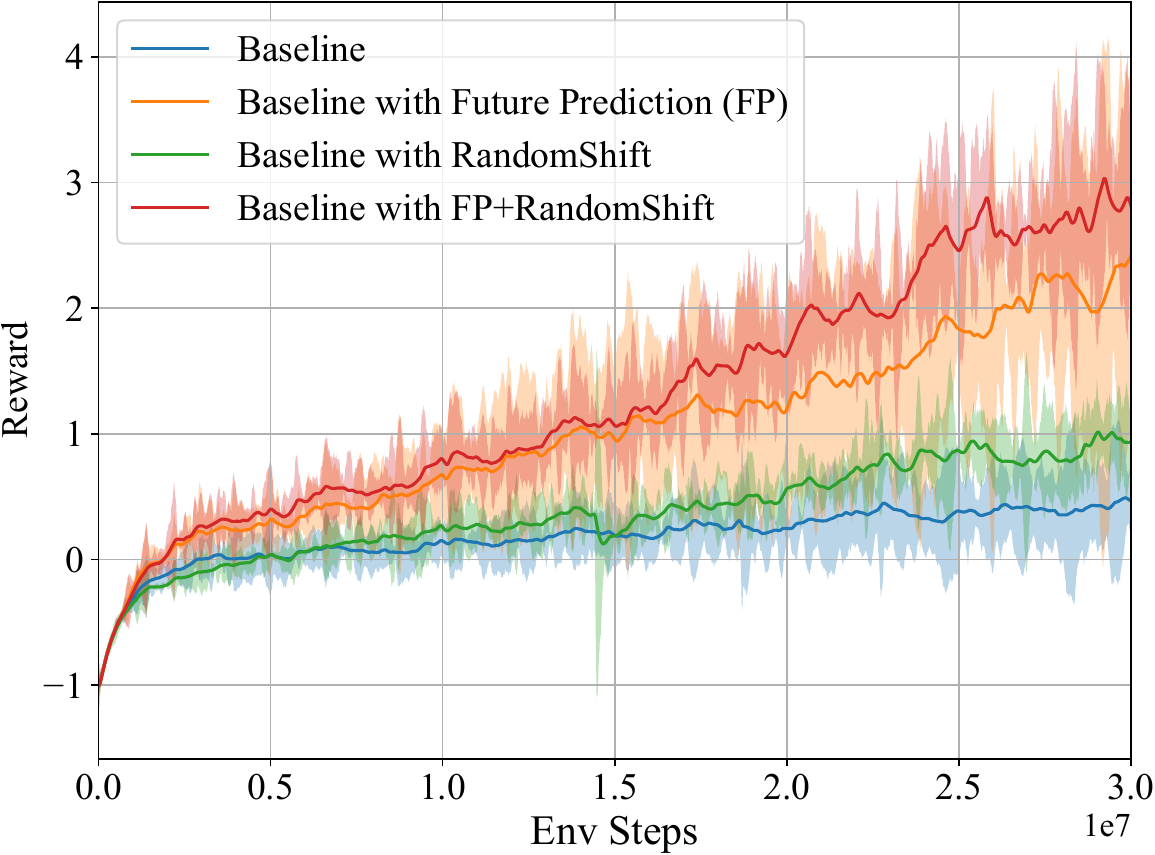}
    \caption{Ablation results of training continuous polices with proposed auxiliary tasks.}
    \label{fig:aug}
    \vspace{-1em}
\end{figure}

\subsubsection{Ablation Study}
To evaluate the effectiveness of two auxiliary tasks in the policy training, ablation experiments with different random seeds are carried out on Gibson benchmark for 30M steps. The results are shown in Fig.~\ref{fig:aug}. Compared to the baseline, the incorporation of either SSL technique accelerates training and enhances performance during the early stages. Combining both with PPO achieves the best results. 

\subsection{Reliability Analysis of Safety Module}
\begin{figure}[!t]
\vspace{+0.5em}
    \centering
    \includegraphics[width=0.75\linewidth]{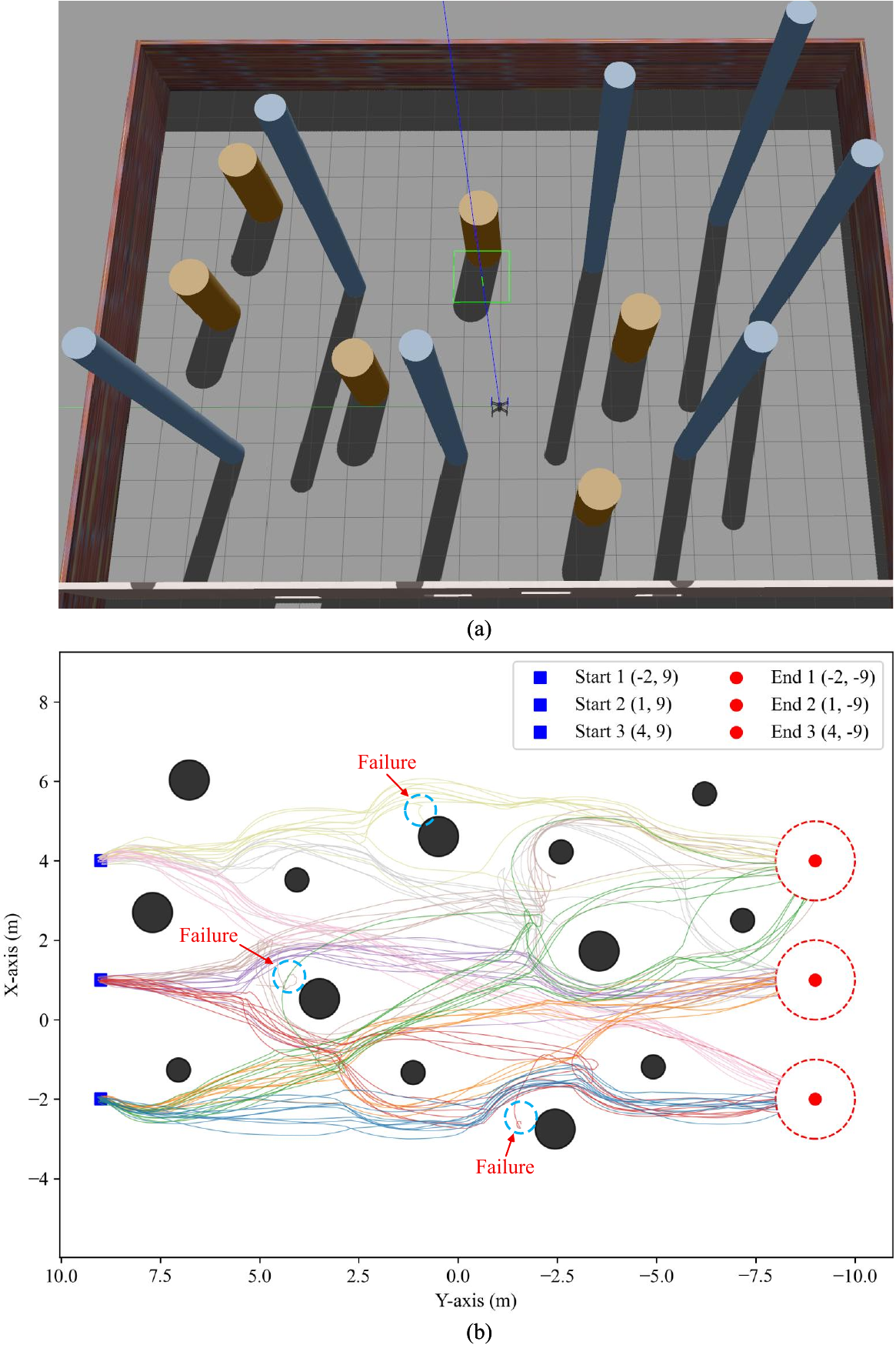}
    \caption{Results of reliability tests for obstacle avoidance. (a) Evaluation environment in Gazebo. (b) Flying trajectories from each start point to the goal point.}
    \label{fig:OA}
    \vspace{-1em}
\end{figure}

Obstacle avoidance is guided by the collision predictor $Q_c$ that is trained together with the navigation policy. To further evaluate its robustness and joint performance with the PX4 flight control system, repeated obstacle avoidance experiments were conducted in Gazebo with pole obstacles. The simulation environment is provided by XTDrone~\cite{xtdrone}, as shown in Fig.~\ref{fig:OA}(a). Gazebo simulator can be integrated with PX4 to closely approximate real-life flight dynamics.

\begin{table}[!b]
\vspace{-1em}
    \caption{Results of obstacle avoidance experiments.\label{table:safe}}
    \centering
    \setlength{\tabcolsep}{3pt}
    \begin{tabular}{cccc}
    \hline Start-End Pair & SPL($\uparrow$) & SR($\uparrow$) & Correction Rate\\
    \hline 
    S1-E1 & 95.8\%  & 100\% & 8.9\% \\
    S1-E2 & 96.6\%  & 100\% & 7.0\% \\
    S1-E3 & 88.7\%  & 100\% & 17.0\% \\
    \hline
    S2-E1 & 83.2\%  & 90\% & 14.6\% \\
    S2-E2 & 96.4\%  & 100\% & 7.2\% \\
    S2-E3 & 76.2\%  & 90\% & 23.9\% \\
    \hline
    S3-E1 & 96.9\%  & 100\% & 6.1\% \\
    S3-E2 & 85.5\%  & 100\% & 19.6\% \\
    S3-E3 & 85.4\%  & 90\% & 10.4\% \\
    \hline
    Average & 89.4\%  & 96.7\% & 12.7\% \\
    \hline
    \end{tabular}
\end{table}

We define a set of start and goal locations, and for each start-goal pair, 10 repeated point-to-point flight trials are conducted. The PX4 controller receives reference commands: angular velocity $v_{ang}$ and forward linear velocity $v_{lin}$. The reference $v_{lin}$ is set as $0.25~m/s$, which matches the maximal speed during ImageNav. The reference $v_{ang}$ is computed based on the yaw deviation between the current orientation and the direction toward the goal. In Fig.~\ref{fig:OA}(b), the blue square indicates the start point, the red circle denotes the end point, and black disks are obstacles. A trial is considered successful if the drone reaches within $1~m$ of the goal without any collision. This setup allows us to evaluate the effectiveness of the proposed action correction mechanism. 

\begin{figure*}[!t]
\vspace{+0.5em}
    \centering
    \includegraphics[width=0.7\linewidth]{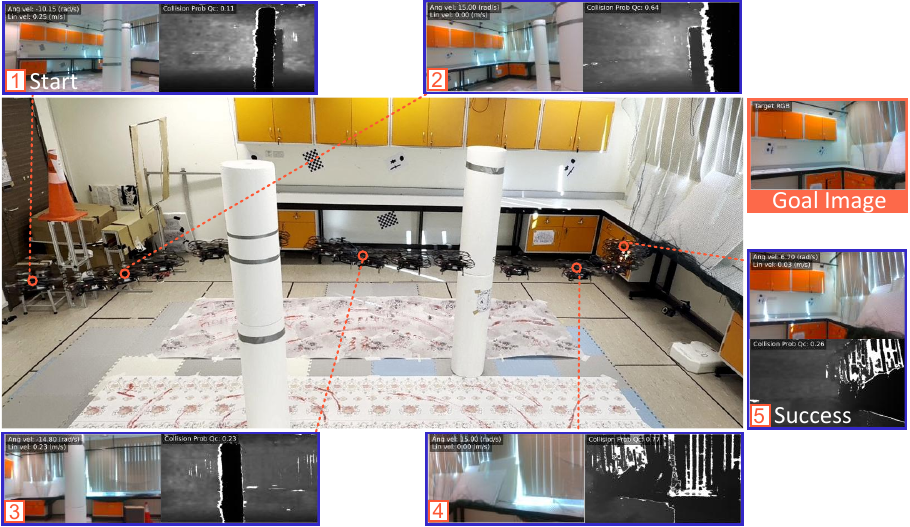}
    \caption{Flight trajectory during the ImageNav task. The image within the red box represents the goal. The intermediate observations, actions, and predicted collision probabilities can be found on the images within the blue boxes.}
    \label{fig:sequence}
    \vspace{-1em}
\end{figure*}

In our experiments, depth perception is limited to $3~\mathrm{m}$, and we denote the normalized depth by $s_t^d$ with $0 \leq s_t^d(i) \leq 1$, corresponding to a real-world range of $[0,3]~\mathrm{m}$. Accordingly, the hyperparameter $\beta=0.3$ in Eq.\eqref{eq:soft} establishes a $0.9~\mathrm{m}$ safety margin, chosen based on the maximal flight speed. Each depth image is cropped to the central $20\%$ horizontal strip, divided into 16 blocks, and the minimum value from each block forms $s_t^d \in \mathbb{R}^{16}$, which is then fed to $Q_c$. In Eq.~\eqref{USL}, the collision probability threshold is set to $d_c=0.5$, corresponding to the Sigmoid function’s decision boundary, and actions are iteratively corrected to ensure $Q_c < d_c$.

\begin{figure}[!t]
  \centering
  \includegraphics[width=0.45\textwidth]{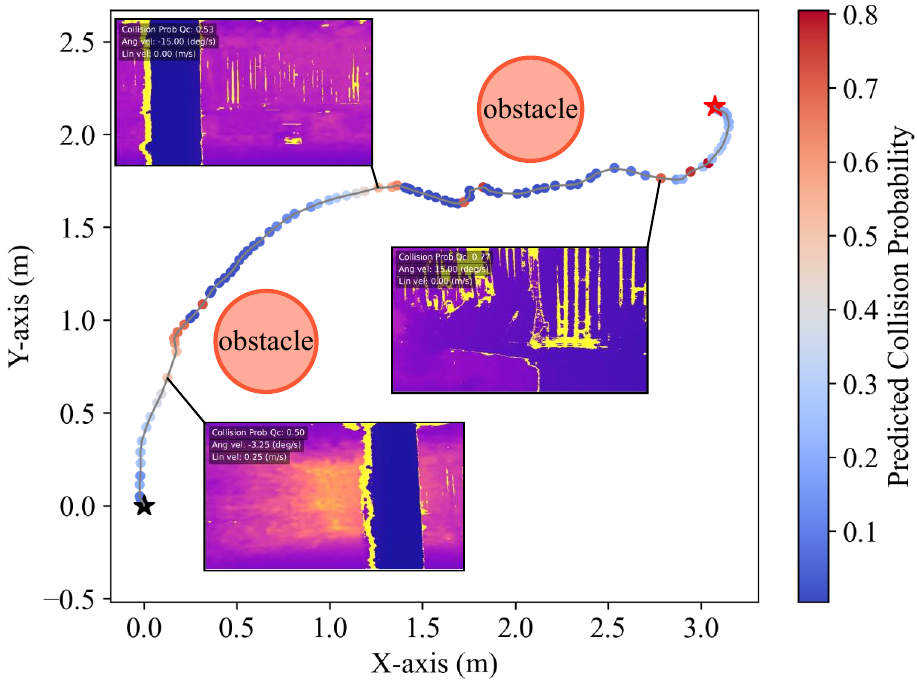}
  \caption{First-person-view depth images at key waypoints for obstacle avoidance. Waypoints are color-coded by their predicted collision probability $Q_c$, with warmer colors indicating higher risk. The trajectory starts at the black star and terminates at the red star.}
  \label{fig:oa}
  \vspace{-1em}
\end{figure}

In our tests, the collision predictor $Q_c$ tends to reduce the collision probability by slowing down $v_{lin}$ and increasing turning speed $v_{ang}$. In some cases, gradient-based correction in Eq.~\eqref{USL} may induce oscillatory motions, as alternating updates to the angular velocity can cause the drone to swing laterally. To mitigate this issue, a more practical way is to apply action correction iteratively at a fixed interval 
\begin{equation}
    \varphi(a_{k}) = a_k-\Delta a = (v_{lin}-\Delta_{lin}, v_{ang}-D \cdot \Delta_{ang}),
\end{equation}
where $D \in \{-1,1\}$ denotes a precomputed safer direction based on the current depth vector $s_t^d$. For a normalized $v_{lin},v_{ang} \in [-1,1]$, we set $\Delta_{lin}=0.3, \Delta_{ang}=0.3$ for both simulations and real-life experiments. To ensure real-time inference, we limit the number of corrections to a maximum of five. If the predicted collision probability is still larger than $d_c$ after five corrections, the forward speed will be reset to 0. Then the drone will rotate in place at a speed of $v_{ang}$ to escape from local dead ends.

The experimental results are presented in Table~\ref{table:safe} and Fig.~\ref{fig:OA}(b). For each start-end pair, we compute the average metrics across 10 trials,  resulting in a total of 90 trajectories. The overall SR is $96.7\%$, SPL is $89.4\%$, and the correction rate is $12.7\%$. Three failure cases are highlighted in Fig.~\ref{fig:OA}(b). The primary cause was inaccurate predictions, which led to a delayed obstacle avoidance. Additionally, when the drone performs an in-place rotation, minor lateral drift will increase the risk of side collisions.

\subsection{Real-life Experiments}
\subsubsection{Hardware Configuration}
We conduct experiments on a drone equipped with an onboard computer and a RealSense D435i camera. The RGB images have a resolution of $848 \times 480$ with a $69^\circ \times 42^\circ$ FoV, while the depth images share the same resolution but feature an $87^\circ \times 58^\circ$ FoV. 
Visual-Inertial Odometry (VIO) is executed onboard, but only the estimated linear velocity and attitude are used for reference tracking control in the PX4 autopilot. The live image stream is transmitted via 2.4 GHz Wi-Fi to an RTX 3060 laptop, with a latency of approximately 10 ms. The velocity control commands \(u = [v_{lin}, v_{ang}]\) are converted from body frame to global frame commands \(u' = [v_{x}, v_{y},v_{ang}]\) using the yaw angle estimated by VIO, and then sent to the flight controller. The planning loop runs at approximately 20 Hz.

In Fig.~\ref{fig:sequence}, real-time velocity commands and the predicted collision probability \(Q_c\) are displayed in the upper-left corner of RGB and depth images. The navigation model explores the environment and approaches areas visually similar to the goal image. Fig.~\ref{fig:oa} shows the flight trajectory along with the predicted collision probabilities at each waypoint.

\section{Discussions and Conclusions}
In this work, we present SIGN, a novel end-to-end RL framework for ImageNav tasks on drones, demonstrating successful sim-to-real transfer. By integrating a visual navigation policy with a safety module, SIGN enables image-goal navigation with reliable collision avoidance. The navigation performance is thoroughly evaluated on three Habitat benchmarks, while the safety module is extensively tested in the Gazebo simulator. Overall, SIGN provides a practical and effective solution for deploying ImageNav on drones.

Our real-world experiments reveal potential failure cases in maze-like environments with visually similar features. When the target image has never been observed, the navigation policy may become stuck in loops, repeatedly exploring the same region without progress.
In future work, we plan to deploy SIGN onboard the Jetson Orin NX platform with a RealSense camera. In preliminary tests, SIGN took 1.1 GB of GPU memory and the inference frequency reached 30 Hz when VIO was running simultaneously. The CPU utilization was about 35\% and the power consumption was 12W.

\bibliographystyle{IEEEtran}
\bibliography{IEEEabrv,reference}

\vfill

\end{document}